\newcommand{\argmin}[1]{\underset{#1}{\operatorname{argmin}}\;}
\newcommand{\Conv}{\mathop{\scalebox{1.5}{\raisebox{-0.2ex}{$\ast$}}}}

 \documentclass[final,5p,times,twocolumn,balance]{elsarticle}

\usepackage{amssymb}
\usepackage{amsthm}
\usepackage{times}
\usepackage{epsfig}
\usepackage{graphicx}
\usepackage{amsmath}
\usepackage{amssymb}
\usepackage{url}
\usepackage{color}
\usepackage{comment}
\usepackage{subfigure}
\usepackage{lineno}
\usepackage{balance}

\journal{Pattern Recognition}

\begin{document}

\begin{frontmatter}



\title{Learn to Model Motion from Blurry Footages}


\author[label1,label2,label3]{Wenbin Li\corref{cor1}}
\author[label3]{Da Chen}
\author[label2]{Zhihan Lv\corref{cor1}}
\author[label4]{Yan Yan}
\author[label5]{Darren Cosker}
\cortext[cor1]{Corresponding author at Department of Computing, Imperial College London, South Kensington London SW7 2AZ, UK.\\ \indent \indent E-mail address: wenbin.li@imperial.ac.uk (Wenbin Li)}
\address[label1]{Department of Computing, Imperial College London, UK}
\address[label2]{Department of Computer Science, University College London, UK}
\address[label3]{Department of Computer Science, University of Bath, UK}
\address[label4]{Department of Information Engineering and Computer Science (DISI), University of Trento, Italy}
\address[label5]{Centre for the Analysis of Motion, Entertainment Research and Applications (CAMERA), University of Bath, UK}

\begin{abstract}


It is difficult to recover the motion field from a real-world footage given a mixture of camera shake and other photometric effects. In this paper we propose a hybrid framework by interleaving a Convolutional Neural Network (CNN) and a traditional optical flow energy. We first conduct a CNN architecture using a novel learnable directional filtering layer. Such layer encodes the angle and distance similarity matrix between blur and camera motion, which is able to enhance the blur features of the camera-shake footages. The proposed CNNs are then integrated into an iterative optical flow framework, which enable the capability of modelling and solving both the blind deconvolution and the optical flow estimation problems simultaneously. Our framework is trained end-to-end on a synthetic dataset and yields competitive precision and performance against the state-of-the-art approaches.
\vspace{1cm}
\end{abstract}

\begin{keyword}
Optical Flow \sep Convolutional Neural Network (CNN) \sep Video/Image Deblurring \sep Directional Filtering



\end{keyword}

\end{frontmatter}


\section{Introduction}

In the image space, the information observed by the dynamical behavior of the object of interest or by the motion of the camera itself is a decisive interpretation for representing natural phenomena. Dense motion, in particular optical flow estimation between a consecutive image pair is the most low-level characterization of such information, which is supposed to estimate a dense field corresponding to the displacement of each pixel. It has become one of the most active fields of computer vision because such characterizations can be extremely embedded into a large number of other higher-level computer vision fields and application domains. Indeed, one can be interested in tracking~\cite{APO,APO_JIFS,tang}, 3D reconstruction~\cite{reflection}, segmentation, as well as the general virtual reality, augmented reality and post-production~\cite{Rotopp2016,tv}.

A typical pipeline of optical flow estimation has been lied on solving a brightness energy with the assistance of patch detection, matching, constrained optimization and interpolation. For many state-of-the-art approaches -- even the precision has reached a reasonable level -- the related applications are still limited by the difficult photometric effects and low performance in runtime. In the recent years, the deep \emph{Convolutional Neural Networks} (CNNs) grows rapidly, which makes a step forward to provide hidden features and end-to-end knowledge representation for many precentral issues \textit{e.g.} motion and texture style \textit{etc}. Such knowledge representation is able to improve the robustness and yields a rapid fashion in the typical optical flow pipeline.

Camera-shake blur is a common photometric effect in the real-world footage, which is often caused by the fast camera motion under a low light condition. Such effect may lead to an invariant blur information for each of the pixel, and may bring extra difficulties into typical optical flow estimation because the basic brightness constancy~\cite{HS} is violated. However, the blur from a daily video footage (24 FPS) can be directionally characterized ~\cite{moBlur}. This observation enables an extra prior to enhance the camera-shake deblurring~\cite{Zhong} and further recover precise optical flow from a blurry images. Such directional prior needs a strict pre-knowledge on the motion direction of the camera which can be obtained by an external sensor~\cite{moBlur}.

\subsection{Contributions}

In this paper, we study the issue of recovering accuracy optical flow from frames of a real-world video footage given a camera-shake blur. The main idea is to learn directional filters, encoded the angle and distance similarity between blur and camera motion. Such filters are further applied to enhance the optical flow estimation. {Our proposed method only relies on the input images, and does not need any other information e.g. ground truth camera motion and blur prior.}


In overview, we propose a novel hybrid approach: \textbf{(1)} we conduct a CNN architecture using a learnable directional filtering layer. Our network is able to extract the blur\&latent features from a blurry image, and further recover the blur kernel within an iterative deconvolutional fashion (Sec.~\ref{sec:cnnDeconv}); \textbf{(2)} we integrate our network into a variational optical flow energy, further optimized within a hybrid coarse-to-fine framework (Sec.~\ref{sec:oflowFramework}).

In the evaluation (Sec.~\ref{sec:eva}), we quantitatively compare our method to four baselines on the synthetic \emph{Ground Truth} (GT) sequences. Those baselines include two blur-oriented optical flow approaches and two other publicly available state-of-the-art methods. We also give quality comparison given real-world blurry footages.

\section{Related Work}

In this section, we will give brief discussion on the related work in specific fields of image deblurring and optical flow estimation.

\subsection{Image Deblurring}
\label{sec:deblurring_related}

{Image blur is a common photometric effect for the daily capture. It is often caused by fast camera movement under a low light condition. Such global blur can be formulated as follows:}

{\setlength\abovedisplayskip{-1mm}
\setlength\belowdisplayskip{-1mm}
\begin{align}
I = k \Conv \ell + n \nonumber
\end{align}
}

{where an observed blurred image $I$ can be represented as a combination of spatial noise $n$ along with a convolution between the latent sharp image $\ell$ and a spatial-invariant blur kernel w.r.t. \emph{Point Spread Function}. To solve the $k$ and $\ell$, a blind deconvolution is normally performed on $I$:}

{\setlength\abovedisplayskip{-1mm}
\setlength\belowdisplayskip{-1mm}
\begin{align}
\argmin{k,\ell}\{ \left \| I - k \Conv  \ell \right \|+\rho(k)\} \nonumber
\end{align}
}

where $\rho$ represents a regularization that penalizes spatial smoothness with a sparsity prior~\cite{FMD}. To solve this ill-posed problem, many approaches rely on additional priors regarding to properties of observed images~\cite{krishnan2011blind,Xu_deblur,panblind,michaeli2014blind,sun2013edge,xiang2012image,yun2011linearized,shao2016regularized}. Pan~\emph{et al.}~\cite{panblind}, for example, propose a blind deconvolution method by taking advantages from the dark channel~\cite{he2011single} regarding to the observation that the dark pixels in the observed image are normally averaged with neighboring pixels along the blur. Krishnan~\emph{et al.}~\cite{krishnan2011blind} introduce a novel scale-invariant regularizer to generate a more stable kernel by fixing the attenuation of high frequencies.

By taking into account the efficient inference, several algorithms~\cite{FMD,xu2010two,Zhong,Shan} are also proposed to solve the deblurring problem. Cho and Lee~\cite{FMD} adopts a predicted edge map as a prior and solve the blind deconvolution energy within a course-to-fine framework. Xu~\emph{et al.}~\cite{xu2010two}, however, discuss a key observation that salient edges do not always help with blur kernel searching. These edges can greatly increase the blur ambiguity in many common scenes. Hence, instead of the use of edge map, they propose an automatic gradient selection scheme to eliminate the ``noisy'' edges for kernel initialization. Furthermore, Zhong~\emph{et al.}~\cite{Zhong} introduce an approach to reduce the noise using a pre-filtering process. Such process preserves the useful image information by reducing the noise along a specific direction.

Both natural image properties based and efficient inference based methods mentioned above are able to provide highly accurate deblurring result for general invariant camera-shake blur. However, these methods often show difficulties given the cases under variant blur. A handful of approaches are proposed to solve such a problem~\cite{gupta2010single,hirsch2011fast,whyte2012non,hu2014joint,khare2016blind}. Gupta~\emph{et al.}~\cite{gupta2010single} propose a \emph{Motion Density Function} to represent the camera motion which is further adopted to recover the spatially varying blur kernel. Hu~\emph{et al.}~\cite{hu2014joint} consider the various depth information of the scene while most of the deblurring methods apply a constant depth for simplicity. They apply an unified layer-based model to jointly estimate the depth and deblurring result from the underlying geometric relationship caused by camera motions.

Since all the methods mentioned above have the specialty along with their limitation, there is no general solution for images blurred by mixed sources, with regard to mixture of fast camera and object movement and scene depth variance. In this case, the image blur is hard to represent by a global model. With the development of Convolutional Neural Network (CNN), some CNN based deblurring methods are proposed to solve such problem. Hradi{\v{s}}~\emph{et al.}~\cite{hradivs2015convolutional} apply a CNN to restore the blurred text documents which is restricted by highly structured data. Xu~\emph{et al.}~\cite{xu2014deep} propose a more general deblurring method. They design a neural network which is guided by traditional deconvolution schemes.

Those mentioned above usually involve a single blurred image as input. There are some hardware assisted methods which are supposed to improve the precision and performance of deblurring~\cite{levin,Joshi,tai08,huimage}. Levin~\emph{et al.}~\cite{levin} propose a uniform method using the known camera arc motion. Such uniformly deblurred image can be estimated by controlling the camera movement along with a parabolic arc. As an extension of this work, Joshi \emph{et al.}~\cite{Joshi} propose to estimate the acceleration and angular velocity of camera by a inertial sensor, \emph{i.e.} gyroscopes and accelerometers. Instead of the highly accurate sensor, Hu~\emph{et al.}~\cite{huimage} introduce a deblurring approach using the smartphone inertial sensors. These methods with extra camera motion information often yield higher performance comparing to those methods only rely on single blurred image as input. However, these methods require complex camera setup and precise calibration.

\subsection{Optical Flow}

\label{sec: of}

\begin{figure*}[t]
\centerline{
\includegraphics[width=1\linewidth]{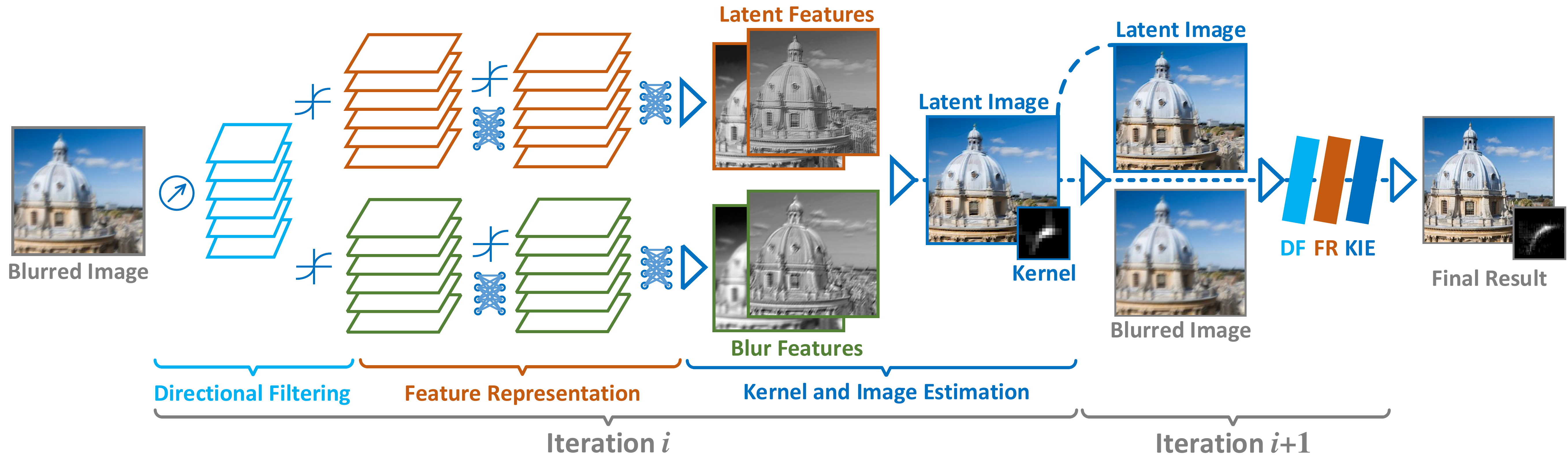}
}
\caption{Our Iterative Deblurring Network. Our network first directionally filters the input image in various directions. The filtered images are then transformed to a learned feature representation for the further kernel and latent image estimation. The resulting latent image together with the blurred one are input to the next iteration until obtaining the final latent image and the blur kernel.}
\label{fig:nnet}
\end{figure*}

Dense motion estimation problem, in particular optical flow, has been widely studied as it can be adopted to many computer vision applications, \emph{e.g.} video segmentation~\cite{grundmann2010efficient}, recognition~\cite{wang2013action} and virtual reality~\cite{vr} \emph{etc.} Many estimation methods have obtained impressive performance in terms of reliability and accuracy showed on the Middlebury~\cite{Middlebury} and Sintel~\cite{Sintel} benchmark. Most of this works are based on the pioneering optical flow method proposed by Horn and Schunck~\cite{HS}. They combine a data term and a smoothness term into an energy function where the former term assumes the certain constancy of the image feature -- typically according to Brightness Constancy Constraint (BCC) -- and the latter term controls how the motion field is varied (such as the Motion Smoothness Constraint). This energy function is then optimized across the entire image to reach the global motion field. This original formula is generally applicable but often limited by many challenges such as large displacement, non-rigid motion, motion boundaries discontinuities, motion blur~\emph{etc.}~\cite{Sintel}. Numbers of extensive works have been proposed to conquer these challenges by introducing additional constraints and more advanced optimization procedure ~\cite{black1996robust,Brox,sun2010layered,xu2012motion,LME,tu2016weighted,ahmad2008human,smokeDa,GT,gt_case,vi_nc}. Brox~\emph{et al.}~\cite{Brox} bring a gradient constancy assumption into the data term in order to reduce the dependency of BCC, and bring a discontinuity-preserving spatio-temporal smoothness constraint to deal with motion discontinuities. Xu~\emph{et al.}~\cite{xu2012motion} propose a novel extended coarse-to-fine (EC2F) refinement framework by taking advantages of feature matching technique. Li \emph{et al.}~\cite{LME} propose to apply laplacian mesh energy to adapt the non-rigid deformation in the scenes.

Moreover, some Neural Network based methods are recently popular. Revaud~\emph{et al.}~\cite{revaud2015epicflow} propose a edge preserving interpolation based on a sparse deep convolutional matching result. The sparse-to-dense interpolation result is then apply to initialize the optimization process for obtaining the final motion field. However, this method strongly relies on the quality of sparse matching where parameters are set manually. Dosovitskiy~\emph{et al.}~\cite{dosovitskiy2015flownet} propose an automatic approach for matching and interpolation. Guiding by a correlation layer, their network can better predict the flow to initialize the refinement. Furthermore, Teney\&Hebert~\cite{teney2016learning} introduce an stand-alone CNN structure for motion estimation requiring less training data. The result, however, is inferior to the state of the art methods.

The presence of blurring features in the scene easily fails the traditional optical flow methods because of the violation to brightness constancy assumption. Only a few of approaches are introduced to settle this problem~\cite{Portz,he2010motion,wulff2014modeling,tu2015estimating,li2013nonrigid}. Portz~\emph{et al.}~\cite{Portz} treat the appearance of each input frame as a parameterized function combining pixel motion and blur motion. The motion clues are then integrated to the data term in energy function. However, it favors the smooth motion field and usually fail at motion boundaries. To solve this problem, Wulff and Black~\cite{wulff2014modeling} treat the motion blur as a function of the layer motion and segmentation determined by a generative model. The optimization is then applied to minimize the pixel error between the input blurred images and synthetic image. Tu~\emph{et al.}~\cite{tu2015estimating} edit the data term using a blur detection based matching method. Their approach is supposed to improve the flow regularization at motion boundaries. Li~\emph{et al.}~\cite{moblur_nc} embed an additional camera motion channel into a hybrid framework in order to obtain the deblurring result and motion estimation result iteratively. {Their method requires a physical motion tracker to obtain the ground truth motion accompanied with the moving camera. Such motion information is supposed to be a hard constraint in the image deblurring step. Besides, their method needs rigid manual tuning for different sequences,  e.g. kernel size, the number of levels of image pyramid, etc.}

In a quick summary, the current methods show extra difficulties to estimate the optical flow from blurry images because the blur may break the photometric properties and further mislead the common regularization. Our proposed method represents the blur image using CNN features which are then used to recover optical flow within a fast optimization framework.

In the following sections, we first discuss our pipeline for recovering the optical flow from a blurred footage (Sec.~\ref{sec:robustEng}). We then introduce the main contributions on our novel CNN based deblurring framework (Sec.~\ref{sec:cnnDeconv}); as well as our hybrid optical flow framework (Sec.~\ref{sec:oflowFramework}) and the evaluation (Sec.~\ref{sec:eva}).

\section{Recover Motion Field from Blurred Footage}
\label{sec:robustEng}

{The typical optical flow framework considers a pair of adjacent images, and follows the \emph{Brightness Constancy} assumption ($E_{data}$) and global smoothness constraint ($E_{sm}$), as follows:}

{\setlength\abovedisplayskip{-1mm}
\setlength\belowdisplayskip{-1mm}
\begin{align}
\centering
E(\textbf{w}) = E_{data}(\textbf{w}) + \gamma E_{sm}(\textbf{w})
\label{eq:Eng}
\end{align}
}

where $I_{1}(\textbf{x})$ and $I_{2}(\textbf{x})$ denote the current frame and its successor respectively. Those observed images can also be represented using a relative latent image and blur kernel, $I_* = k_*\Conv \ell_*$, $\{I_{*}, \ell_*:\Omega \subset \mathbb{R}^3\}$. The optical flow field, denoted by $\textbf{w} = (u,v)^{T}$ can be obtained by solving this functional.

{However, given such a pair of blurred images, the blur information may damage image structure and further violate the basic \emph{Brightness Constancy} assumption of optical flow estimation. Those large number of outliers would lead to uncertain errors to energy optimization. The straight forward solution is to remove the blur before performing the optical flow estimation. The deblurring process may sharpen the images but still permanently change the pixel intensity and further bring unpredictable artifacts. The alternative is to match un-uniform blur~\cite{moblur_nc,Portz} between the input images:}


{\setlength\abovedisplayskip{-1mm}
\setlength\belowdisplayskip{-1mm}
\begin{align}
B_1 = k_2\Conv I_1 \approx k_2 \Conv k_1 \Conv \ell_1 \nonumber\\
B_2 = k_1\Conv I_2 \approx k_1 \Conv k_2 \Conv \ell_2
\label{eq:uniformImg}
\end{align}
}

where we have uniform blur images $B_1$ and $B_2$ which is supposed to use in the \emph{Blur Brightness} and \emph{Blur Gradient Constancy} terms:

{\setlength\abovedisplayskip{-1mm}
\setlength\belowdisplayskip{-1mm}
\begin{align}
E_{data}(\textbf{w}) &= \underbrace{\int_{\Omega} \phi ( \left \| B_{2}(\textbf{x}+\textbf{w}) - B_{1}(\textbf{x})\right \|^{2})d\textbf{x}}_{Blur~Brightness~Constancy}\nonumber\\
&+ \alpha \underbrace{\int_{\Omega} \phi (\left \| \nabla B_{2}(\textbf{x}+\textbf{w}) - \nabla B_{1}(\textbf{x})\right \|^{2})d\textbf{x}}_{Blur~Gradient~Constancy}
\end{align}
}

where $\nabla = (\partial_{xx},\partial_{yy})^{T}$ denotes a spatial gradient and $\alpha\in[0,1]$ presents a linear weight. The smoothness term regularizes the global flow variation as follows:

{\setlength\abovedisplayskip{-1mm}
\setlength\belowdisplayskip{-1mm}
\begin{eqnarray}
E_{sm}(\textbf{w}) = \int_{\Omega} \phi(\left \| \nabla u \right \|^{2} + \left \| \nabla v \right \|^{2})d\textbf{x}
\end{eqnarray}
}

where Lorentzian regularization $\phi(s)=log(1+s^2/2\epsilon^{2})$ is applied to preserve motion boundaries. {The un-uniform blur matching is supposed to protect the color properties of the images, as well as further keep color correlation and consistency across the input images. In Table~\ref{tab:eva_self}, we quantitatively evaluate how the blur matching significantly improves the flow precision.}

In the following sections, we present our CNNs based approach which consists two stacked modules: \textbf{(1)} A layered network for blind deconvolution; \textbf{(2)} An iterative optical flow framework.

\section{A Layered Network for Blind Deconvolution}
\label{sec:cnnDeconv}

\begin{figure*}[t]
\centerline{
\includegraphics[width=0.93\linewidth]{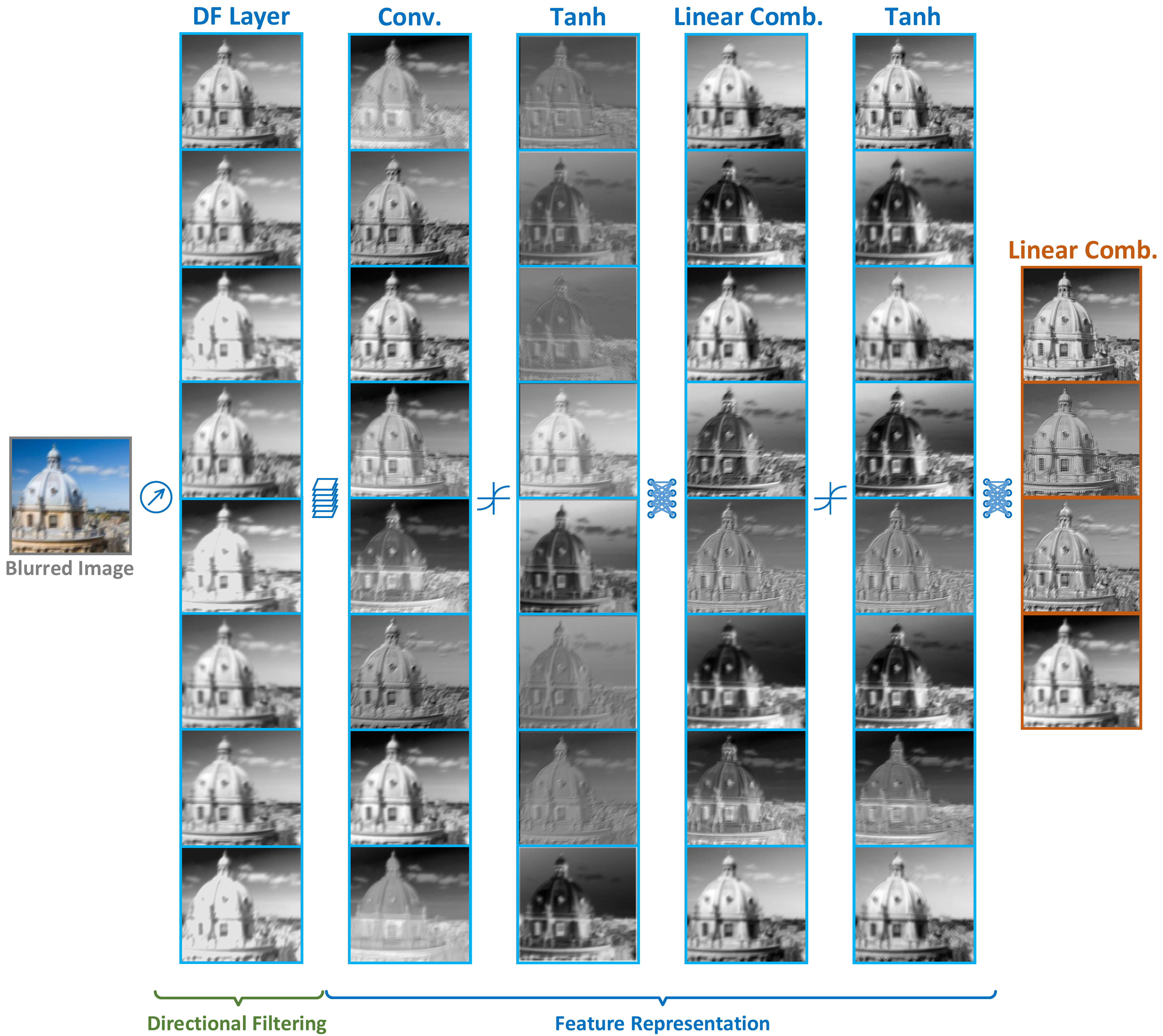}
}
\caption{Intermediary Visualization for Layers. In the implementation, our network consists six layers, \textbf{From Left to Right}: directional filtering, convolution filtering, tanh nonlinearity, linear combination on hidden layers, tanh nonlinearity and linear combination to four feature representations in particular two sharp images and two blurry ones.}
\label{fig:layers}
\end{figure*}

As show in Fig.~\ref{fig:nnet}, we propose an $n$-iteration coarse-to-fine blind deconvolution module which takes into account a trainable convolutional neural network. For each iteration, the input images are operated with the following processes:

\subsection{Directional Filtering}

{The blur from daily photography may be highly nonlinear and hard to predict. This, however, can be parameterized as a near linear form if it is from a daily video footage captured at ordinary frame rate (24 Hz). In this context, direction filters~\cite{Zhong,moblur_nc} may be effective to regularize the blur within deconvolution operation. A common form reads:}

{\setlength\abovedisplayskip{-1mm}
\setlength\belowdisplayskip{-1mm}
\begin{align}
f(\omega) \circ I(\textbf{x}) =\int_\textbf{x}\int_t \kappa G(t)I(\textbf{x}+t\omega)d\textbf{x}dt
\label{eq:filtering}
\end{align}
}

where $\textbf{x}=(x,y)^T$ represents a pixel location and $G$ denotes a Gaussian kernel and $\omega=(\cos\theta,\sin\theta)^T$ controls the filtering direction. The filter is further normalized by $\kappa=\left ( \int G(t)dt\right )^{-1}$. Such directional filter is able to remove the general noise but does not affect the signal along the orthogonal direction. Given a ground truth blur direction $\omega$, the filtered image $\widetilde{I}^\omega = f(\omega+\pi/2) \circ I$, may destroy the color properties but is supposed to enhance the useful blur information.

In our network, we propose a novel \emph{Directional Filtering} (DF) layer which calculates a new group of image representation by applying a directional filter across different directions. This process aims to remove the spatial noise while preserve the blur information. We therefore model the first filtering layer using shared weights across all the locations. Our filters read:

{\setlength\abovedisplayskip{-1mm}
\setlength\belowdisplayskip{-1mm}
\begin{align}\label{eq:dfLayer}
  \widetilde{I}_i = \varphi_i f(i\pi/16)\circ I
\end{align}
}

where $f(\omega)$ denotes the directional filter and $\varphi_i$ weights the strongness for the specific direction of the filtering. $i$ is the number of the direction sampling. In our implementation, we uniformly select $k+1$ directions $i=\{0, 1,\cdots, k\}$ within $\pi/2$. After the directional filtering, we further construct the deep feature representation for the images.

\subsection{Feature Representation}
Similar to Schuler~\emph{et al.}~\cite{learndeblur}, our network does not predict the latent image directly but perform a \emph{Feature Representation} which computes gradient image representations and preliminary estimates for the further kernel and latent image estimation. Our scheme is supposed to extract the features from a subset of pixels. The local feature information is then integrated by a global combination. Such heuristic strategy can greatly shrink the number of parameters for optimization.

To extract our \emph{Feature Representation}, we adopt a sub-network with three layers, in particular convolution, nonlinearity and linear combination respectively. We first apply a group of \emph{Convolutional Filters} (CFs) onto the denoised (directional filtered) images, then transform the values using tanh function. In this case, the resulting features are further combined linearly for a new representation as follows:

{\setlength\abovedisplayskip{-1mm}
\setlength\belowdisplayskip{-1mm}
\begin{align}\label{eq:convLayer}
  \widehat{I}_i=\sum_{j}\lambda_{ij}\,\texttt{tanh}(\mathcal{C}_j \Conv \widetilde{I})\nonumber\\
  \widehat{\ell}_i=\sum_{j}\delta_{ij}\,\texttt{tanh}(\mathcal{C}_j \Conv \widetilde{I})
\end{align}
}

where $\mathcal{C}_*$ denotes a set of \emph{Convolutional Filters} that are shared between the $\widehat{I}_*$ and $\widehat{\ell}_*$. $\texttt{tanh}(\cdot)$ presents the nonlinearity while the $\lambda_{**}$ and $\delta_{**}$ weight the linear combinations for the $\widehat{I}_*$ and $\widehat{\ell}_*$ respectively. Fig.~\ref{fig:layers} shows the intermediary features for each layer of our network. Note that we stack the tanh and linear combination layers twice after the convolutional layer for proper level of nonlinearity~\cite{learndeblur}. In practice, those layers can be further stacked for difficult cases i.e. strong blur and noise~\cite{xu2014deep}.

After those layers, we obtain a new feature representations for $\widehat{I}_*$ and $\widehat{\ell}_*$ respectively. Those featured images are then used to estimate the latent image and kernel.

\subsection{Kernel and Latent Image Estimation}

Once we have the current feature representation $\widehat{I}_*$ and $\widehat{\ell}_*$, a variation of Cho\&Lee~\cite{FMD,moblur_nc} is adopted for the \emph{Kernel and Latent Image Estimation}. Our method consists two steps: \textbf{(1)} Given $\widehat{I}_*$ and $\widehat{\ell}_*$, we calculate their gradient maps $\Delta = \{\partial_x,\partial_y\}$ along the horizontal and vertical directions, which is capable of further preserving the high frequency information i.e. edges and image structure. \textbf{(2)} Those resulting gradient maps $\Delta \widehat{I}$ and $\Delta \widehat{\ell}$ are then used to optimize the energy:

{\setlength\abovedisplayskip{-1mm}
\setlength\belowdisplayskip{-1mm}
\begin{align}
\widehat{k} = \argmin{\widehat{k}} \sum_{\widehat{I}_*,\widehat{\ell}_*}\tau_* \left \| \widehat{I}_* - \widehat{k}\Conv \widehat{\ell}_*\right \|^2+\beta_{k}\left \| \widehat{k} \right \|^2\nonumber\\
(\widehat{I}_*,\widehat{\ell}_*) \in \{(\partial_x \widehat{I}, \partial_x \widehat{\ell}), (\partial_y \widehat{I}, \partial_y \widehat{\ell}),(\partial_{xx} \widehat{I}, \partial_{xx} \widehat{\ell}), \nonumber\\
(\partial_{yy} \widehat{I}, \partial_{yy} \widehat{\ell}),(\partial_{xy} \widehat{I}, (\partial_x\partial_y +\partial_y \partial_x)\widehat{\ell}/2)\}
\label{eq:FMeng}
\end{align}
}

\begin{figure}[t]
\centerline{
\includegraphics[width=1\linewidth]{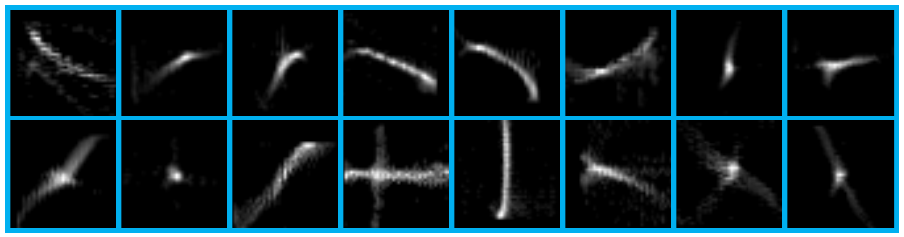}
}
\caption{Kernels for Training. To train our network, 16 kernels are estimated from a real-world footage~\cite{moblur_nc}. We then generate 10 variations for each of kernel by rotating $n\pi/20$ radians. Those kernels are applied to the sharp training images after resized to various sizes.}
\label{fig:kernels}
\end{figure}

where $\tau_*$ linearly weights the derivatives in either directions while $\beta_k$ presents a weight for Tikhonov regularization on the kernel. Here both initial $\widehat{I}$ and $\widehat{\ell}$ are from our feature representations. The proposed energy function above is highly nonlinear. which is minimized by following an iterative numerical scheme from \cite{FMD,moblur_nc}. The resulting pre-optimal kernel $\widehat{k}$ can be used to estimate the latent image $\ell$ within a \emph{Non-blind Deconvolution} process:

{\setlength\abovedisplayskip{-1mm}
\setlength\belowdisplayskip{-1mm}
\begin{align}
\ell = \argmin{\ell} \sum_{i}\left \| \widehat{I}_i - \widehat{k}\Conv \ell\right \|^2+\beta_\ell\left \| \widehat{I}_i \right \|^2
\label{eq:nonblindDeconv}
\end{align}
}

By minimizing the energy function above, we obtain the latent image $\ell$. Depending on the desired quality of the final deblurring, $\ell$ can be stacked to the blurred image $I$ along the third dimension. Such stacked image is input to our network and run through layers iteratively until obtaining the final $\ell$ and $k$. In this case, all the learned filters of our network have three dimensions.

In summary, our network only regularizes the the free parameters in \emph{Directional Filtering} and \emph{Feature Representation} but fixes the hyper-parameter in \emph{Kernel and Image Estimation}. In this case, similar to~\cite{learndeblur}, our learning model sticks on learning filters with a limited receptive field instead of the full dimensionality of the input blurred images.

\subsection{Parameter Training}
\label{sec:train}

\begin{figure}[t]
\centerline{
\includegraphics[width=1\linewidth]{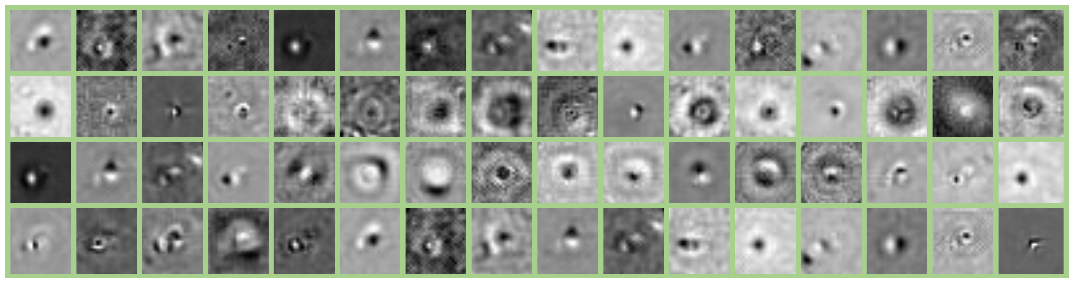}
}
\caption{Features Learnt from Our Network.}
\label{fig:feature}
\end{figure}

Similar to the traditional approaches, we synthesize pairs of latent and blurred images to train our network. We randomly sample 1,000 images (cropped to 480$\times$480 pix.) from the recent large-scale datasets~\cite{MIFDB16} which contains around 34,000 feature-rich sharp images from three synthetic scenes. To synthesize associated blurred images, we first adopt 16 kernels from a real-world footage~\cite{moblur_nc}. As shown in Fig.~\ref{fig:kernels}, those kernels are near linear. For each of these selected kernels, 10 variations are generated by rotating for $n\pi/20$ radians. In this case, we obtain 160 distinguishing kernel variations. We randomly resize each of those kernels into the size between $15\times15$ pixel and $35\times35$ pixel; then apply them to the selected sharp image respectively. After this process, we obtain 160,000 pairs of training images. During the training, we randomly add either Gaussian (0.01) or Salt\&Pepper (0.15) noise.

For obtaining a proper result, we may perform $N$ iterations on our network, which leads to a large number of parameters for training. Here we follow the stage based training strategy from~\cite{learndeblur}. We start with the first iteration by using the $\mathcal{L}_2$ loss function onto the ground truth and estimated results. We then fix the parameters from previous iteration but only update the ones of the next iteration until the last iteration. This training process is supposed to be more efficient against the end-to-end strategy because it limits the number of updated parameters for different training stages. In practice, we adopt a fixed learning rate (0.01)
and a decay rate (0.95).

\begin{figure}[t]
\centerline{
\includegraphics[width=0.85\linewidth]{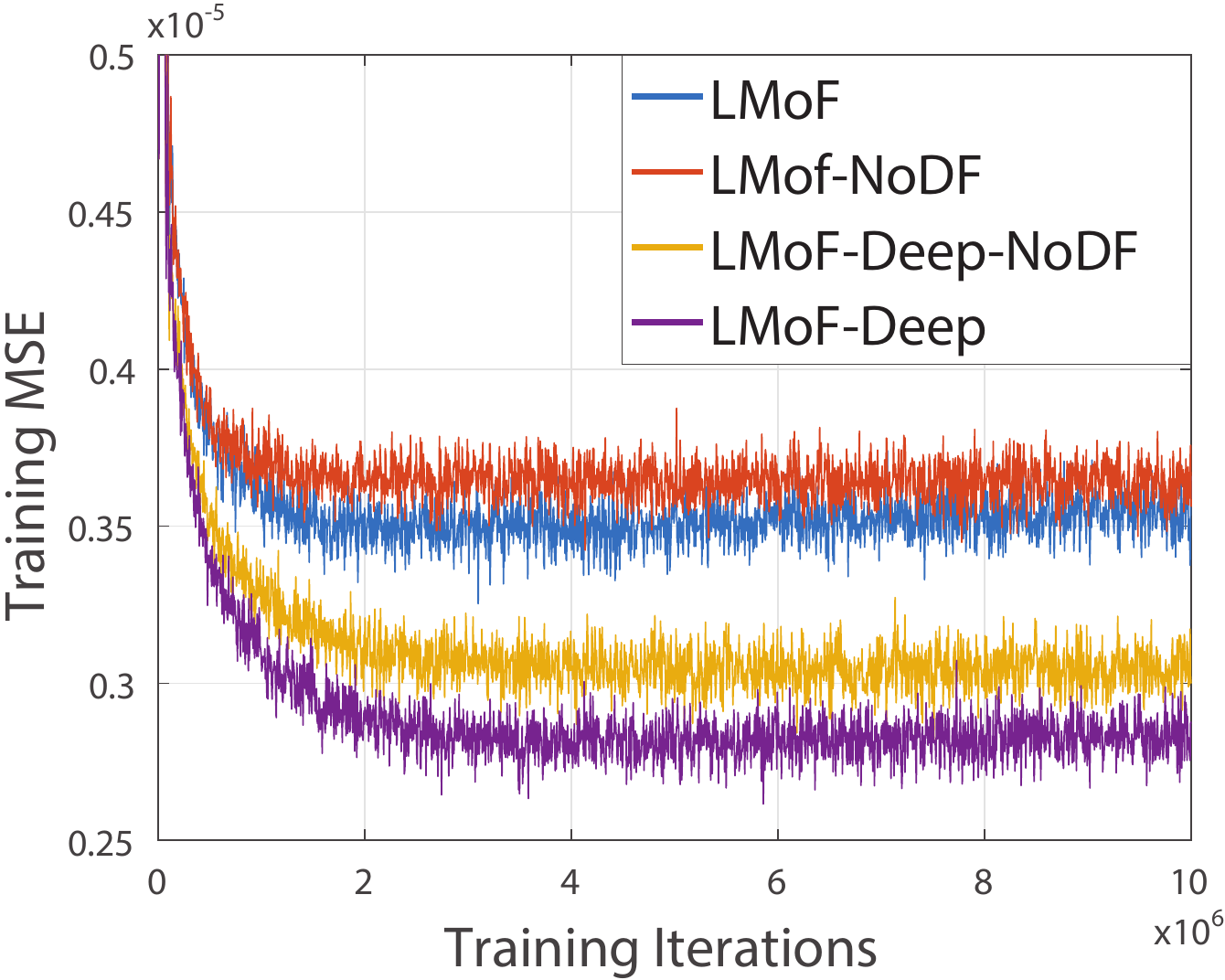}
}
\caption{Training Performance on Our Network. The MSE metric is measured during the training on different architectures of our network (Sec.~\ref{sec:train}). More number of hidden layers, DFs and CFs can improve the general performance.}
\label{fig:trainErr}
\end{figure}

Fig.~\ref{fig:trainErr} illustrates the training performance on four variations of our approach i.e. \emph{LMoF}, \emph{LMoF-Deep}, \emph{LMoF-NoDF} and \emph{LMoF-Deep-NoDF}. Here \emph{LMoF} denotes our method using a network of 8 DFs, 8 CFs, 8 hidden layers and linear combination while {\emph{LMoF-Deep} presents an enhanced version by using a deeper network of 24 DFs, 48 CFs, 30 hidden layers and linear combination.} \emph{LMoF-NoDF} and \emph{LMoF-Deep-NoDF} are the related versions \textbf{without} the DF layer. \emph{LMoF} and \emph{LMoF-Deep} outperform the related versions without the DF layer. {It is worthy noting that with a larger number of DFs, CFs, hidden layers and network iterations, the deblurring quality can be greatly improved.

The experiments in table~\ref{tab:eva} (LMoF versus LMoF-NoDF) illustrate the final optical flow precision is improved by around 20\% for all trials. On the other hand, the deeper version (LMoF-Deep) shows similar comparative measure (LMoF-Deep also gives lower training error against LMoF-Deep-NoDF; and LMoF-Deep gives general the best precision for most of trials in Table~\ref{tab:eva}) We believe that those improvement on training/results is not about the overfitting. However, the number of the network iterations significantly affects the computational speed. In this context, we run three iterations for each of our implementations to balance the general performance with precision.}

\begin{table}[t]
\centerline{
\begin{tabular}{l}
\hline
\hspace{5 pt}\textbf{Algorithm 1}: CNN based Optical Flow Framework\\
\hline
\hspace{5 pt}\textbf{Input} \hspace{6 pt} :~A blurred image pair $I_1$, $I_2$\\
\hspace{5 pt}\textbf{Output}~:~Optical flow field $\textbf{w}$\vspace{1pt}\\
\hspace{5 pt}1:\hspace{5 pt} \emph{Construct an $N$-level top-down image pyramid} \\
\hspace{5 pt}2:\hspace{5 pt} Level index $i = \{0, 1, \dots, N\}$\\
\hspace{5 pt}3:\hspace{5 pt} $\ell^i_1 \leftarrow I^i_1$, $\ell^i_2 \leftarrow I^i_2$, $\textbf{w}^i\leftarrow (0,0)^T$\vspace{1pt}\\
\hspace{5 pt}4:\hspace{5 pt} \textbf{for} \emph{coarse to fine} \textbf{do}\\
\hspace{5 pt}5:\hspace{30 pt} $i \leftarrow i+1$\\
\hspace{5 pt}6:\hspace{30 pt} $\ell^i_{\{1,2\}}$, $I^i_{\{1,2\}}$ and $\textbf{w}^i$ resized to the $i$th scale\\
\hspace{5 pt}7:\hspace{30 pt} \textbf{foreach} $* \in \{1,2\}$ \textbf{do}\\
\hspace{5 pt}8:\hspace{55 pt} $\widehat{\ell}^i_*, \widehat{I}^i_* \leftarrow$ \texttt{CNNFeatureNet} ( $\ell^i_*, I^i_*$ )\\
\hspace{5 pt}9:\hspace{55 pt} $k^i_* \leftarrow$ $\texttt{NonBDeconv}^{1st}$ ( $\widehat{\ell}^i_*, \widehat{I}^i_*$ )\\
11:\hspace{55 pt} $\ell^i_* \leftarrow$ $\texttt{NonBDeconv}^{2nd}$( $k^i_*, I^i_*$ )\\
12:\hspace{30 pt} \textbf{endfor}\\
13:\hspace{30 pt} $B^i_1 \leftarrow I^i_1 \Conv k^i_2$, $B^i_2 \leftarrow I^i_2 \Conv k^i_1$\\
14:\hspace{30 pt} $d\textbf{w}^i\leftarrow$ \texttt{EnergyOpt} ( $B^i_1, B^i_2, \textbf{w}^i$ )\\
15:\hspace{30 pt} $\textbf{w}^i\leftarrow \textbf{w}^i+d\textbf{w}^i$\\
16:\hspace{5 pt} \textbf{endfor}\\
\hline
\end{tabular}
\vspace{-3mm}
}
\label{alg}
\end{table}

In the next section, we embed our proposed network into an optical flow framework.

\section{An Iterative Optical Flow Framework}
\label{sec:oflowFramework}

Algorithm~1 sketches the proposed \emph{CNN based Optical Flow Framework} which interleaves our layered network and an iterative optical flow optimization.

Within our framework, the input images $I_{\{1,2\}}$ are first resized into a coarse-to-fine (top-down) pyramid. On each pyramidical level $i$, \textbf{(1)} the resized images $I^i_{\{1,2\}}$ are input into our \emph{Layered Network for Blind Deconvolution} (Sec.~\ref{sec:cnnDeconv}) which yields intermediary latent images $\ell^i_*$ and kernels $k^i_*$. \textbf{(2)} Those information is then used to generate uniform \emph{Motion Energy for Blurred Images} (Sec.~\ref{sec:robustEng}). \textbf{(3)} Such blurred energy is optimized for the incremental optical flow field $d\textbf{w}^i$ (Sec.~\ref{sec:mini}). Finally, those parameters $\ell^i_*$ and $\textbf{w}^i$ are then propagated to the next level until convergence. {Note that our framework is not a simple combination of image deblurring and optical flow estimation. Our \emph{Layered Network for Blind Deconvolution} is deeply embedded (Per-level) into every level of image pyramid. And the following blur matching step (step 13, Algorithm 1) further preserves brightness constancy. In this case, the CNN based deblurring process is automatically optimized against the size of image (different levels of image pyramid). Table~\ref{tab:eva_self} quantifies the advantage of our Per-level strategy given the ground truth dataset.}

In the next subsection, we introduce the our energy optimization scheme in details.

\subsection{Energy Optimization}
\label{sec:mini}

\begin{figure}[t]
\centerline{
\includegraphics[width=1\linewidth]{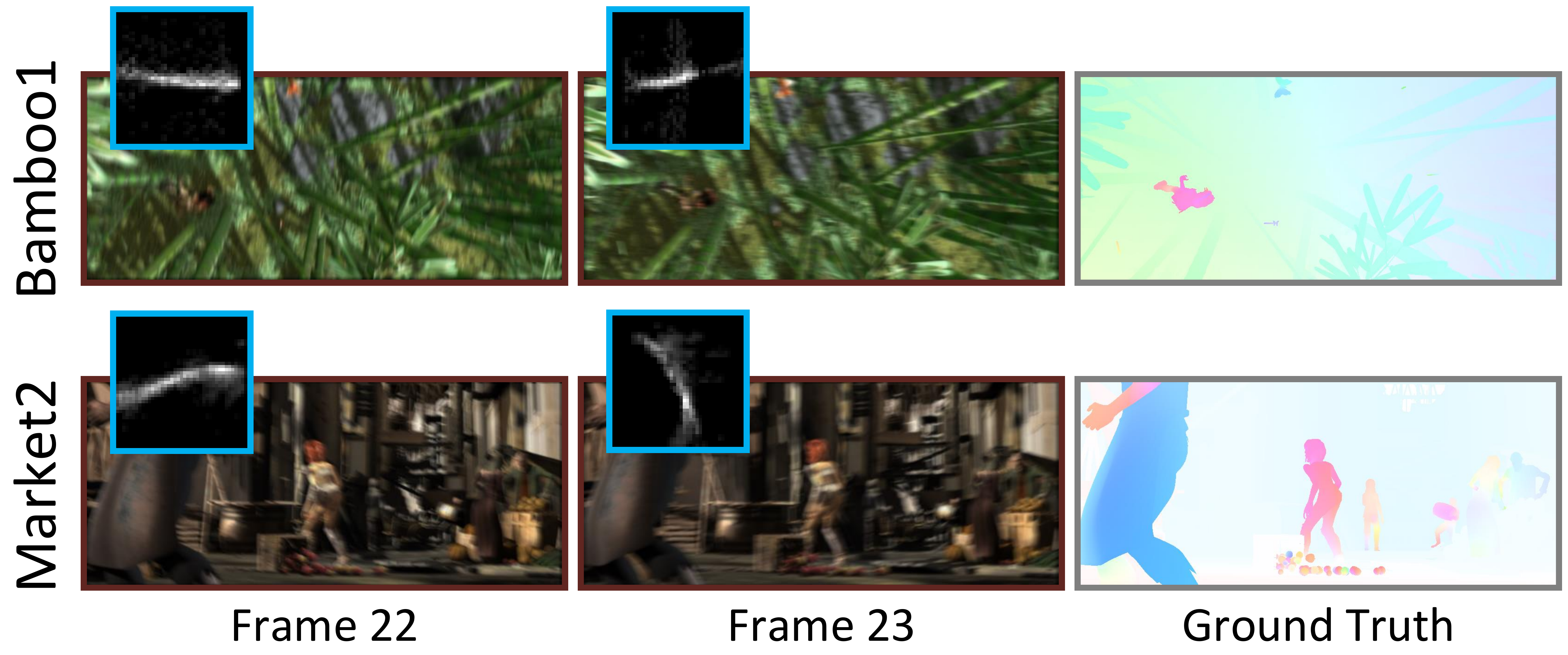}
}
\caption{Synthetic Ground Truth Sequences. Our extra ground truth sequences generated by applying real-world blur kernels onto the selected frames of Sintel sequences Bamboo1 and Market2.}
\label{fig:gt}
\end{figure}

\begin{table*}[t]
\centerline{
    \begin{tabular}{|c||c|cc|cc|cc|cc||cc|cc|}
    \hline
        & \multicolumn{1}{c|}{\bfseries{}}
        & \multicolumn{2}{c|}{\bfseries{Grove2}}
        & \multicolumn{2}{c|}{\bfseries{Hydrangea}}
        & \multicolumn{2}{c|}{\bfseries{Rub.Whale}}
        & \multicolumn{2}{c||}{\bfseries{Urban2}}
        & \multicolumn{2}{c|}{\bfseries{Bamboo1}}
        & \multicolumn{2}{c|}{\bfseries{Market2}}\\
        \bfseries{Baselines}
        & \bfseries{Time}
        & \multicolumn{1}{c}{AEE}
        & \multicolumn{1}{c|}{AAE}
        & \multicolumn{1}{c}{AEE}
        & \multicolumn{1}{c|}{AAE}
        & \multicolumn{1}{c}{AEE}
        & \multicolumn{1}{c|}{AAE}
        & \multicolumn{1}{c}{AEE}
        & \multicolumn{1}{c||}{AAE}
        & \multicolumn{1}{c}{AEE}
        & \multicolumn{1}{c|}{AAE}
        & \multicolumn{1}{c}{AEE}
        & \multicolumn{1}{c|}{AAE}\\ 
        \hline
        {\textbf{LMoF-Deep}} & 35 & $\textbf{0.39}^1$ & $\textbf{1.40}^1$ & $\textbf{0.49}^1$ & $\textbf{1.18}^1$ & $\textbf{0.53}^1$ & $\textbf{2.31}^1$ & $\textbf{0.94}^1$ & $\textbf{2.11}^1$ & $\textbf{1.01}^1$ & $\textbf{3.12}^1$ & $\textbf{3.28}^1$ & $8.21^2$\\
        \textbf{LMoF} & 29 & $0.62^3$ & $2.12^2$ & $0.88^3$ & $1.90^2$ & $0.86^3$ & $3.31^2$ & $1.28^2$ & $2.99^3$ & $2.16$ & $4.89^3$ & $4.23^3$ & $8.90$\\
        \textbf{LMoF-NoDF} & 27 & 0.71 & 2.78 & 0.96 & $2.08^3$ & 0.98 & 3.88 & 1.42 & 3.12 & 2.46 & 8.61 & 4.54 & 8.98\\
        \textbf{Li}~\emph{et al.}~\cite{moblur_nc} & 39 & $0.47^2$ & $2.34^3$ & $0.67^2$ & 2.19 & $0.62^2$ & $3.67^3$ & $1.36^3$ & $2.87^2$ & $1.60^2$ & $4.19^2$ & $4.31$ & $8.72^3$\\
        \textbf{Portz}~\emph{et al.}~\cite{Portz} & 79 & 1.14 & 4.11 & 2.62 & 3.55 & 3.12 & 8.18 & 3.44 & 5.10 & 2.32 & 9.02 & 4.68 & 8.91\\
        \textbf{Brox}~\emph{et al.}~\cite{Brox} & 22 & 1.24 & 4.53 & 2.26 & 3.47 & 2.44 & 7.98 & 2.92 & 4.60 & 4.86 & 5.69 & 6.96 & 10.18\\
        \textbf{MDP}~\cite{xu2012motion} & 422 & 1.06 & 3.46 & 3.40 & 3.55 & 3.70 & 8.21 & 5.62 & 6.82 & 2.97 & 10.54 & 5.88 & 9.59\\
        \hline
        {\textbf{FlowNetS}}~\cite{dosovitskiy2015flownet} & 0.09 & 1.31 & 4.48 & 1.78 & 3.37 & 1.20 & 6.75 & 2.55 & 4.24 & $2.00^{3}$ & 7.21 & $4.01^2$ & $\textbf{8.18}^1$\\
        {\textbf{FlowNetC}}~\cite{dosovitskiy2015flownet} & 0.12 & 1.43 & 4.80 & 2.49 & 3.96 & 1.87 & 7.15 & 3.60 & 5.55 & 1.96 & 6.51 & 4.18 & 7.98\\
        {\textbf{Teney\&Hebert}}~\cite{teney2016learning} & 7 & 0.78 & 3.21 & 0.91 & 2.78 & 0.88 & 5.74 & 1.97 & 4.10 & 2.93 & 6.33 & 5.49 & 9.09\\
        \hline
    \end{tabular}
    }
    \vspace{2mm}
    \caption{Quantitative Measure on GT dataset (Li~\emph{et al.}'s benchmark + our customized Sintel). Our method (LMoF-Deep, LMoF and LMoF-NoDF) is compared to the other baselines on the metrics of Average Endpoint Error (AEE), Average Angle Error (AAE) and Average Time Consumption (in second).}
    \label{tab:eva}
\end{table*}

\begin{figure*}[t]
\centerline{
\includegraphics[width=0.95\linewidth]{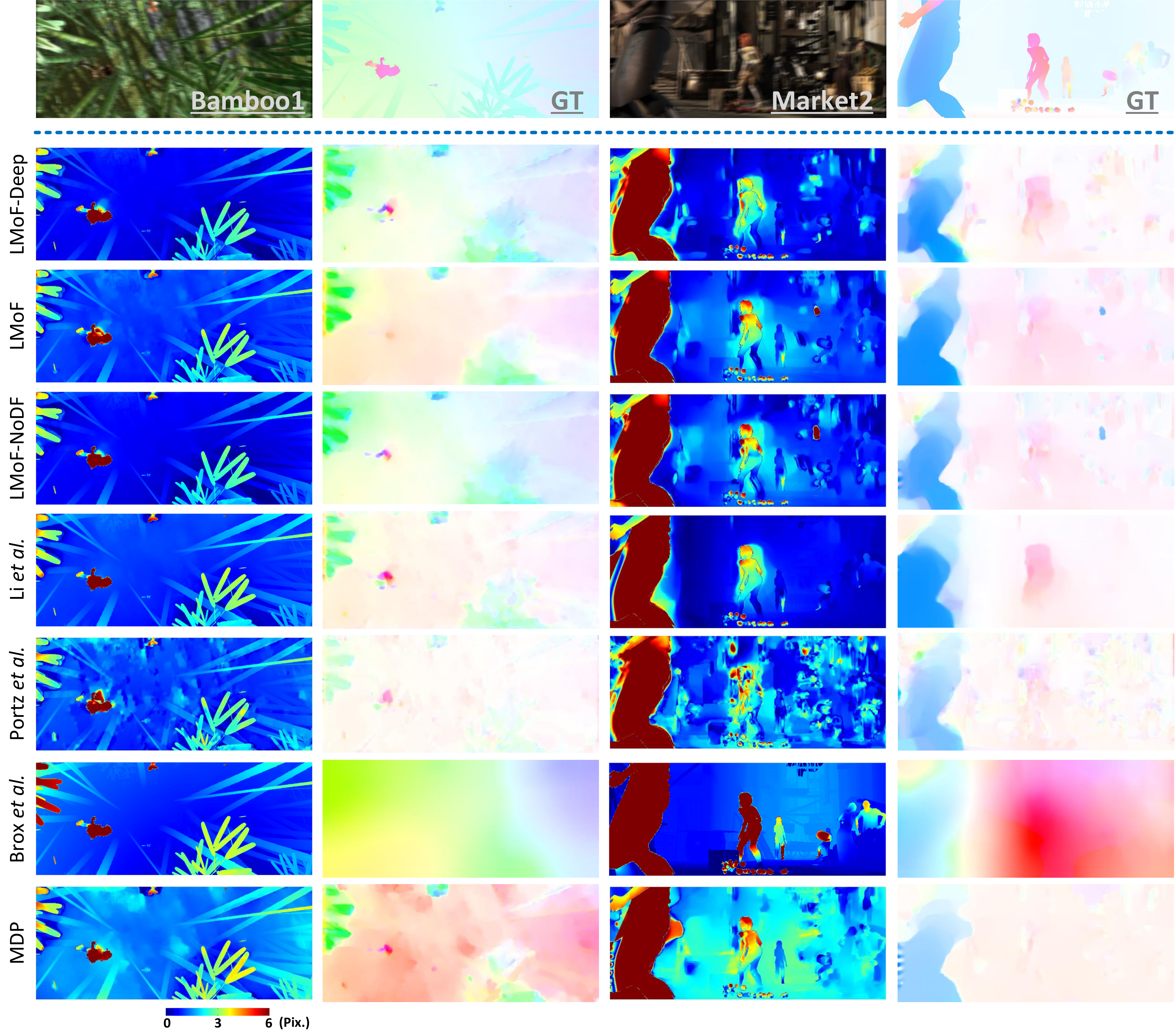}
}
\caption{Visual Comparison on \emph{Bamboo1} and \emph{Market2}. \textbf{First Row:} the blurry images and the GT flow fields. From left to right, the \textbf{First} and \textbf{Third Columns} are the error map comparing to the GT flow fields. The \textbf{Second} and \textbf{Last Columns} are the flow fields of each baselines.}
\label{fig:synErr}
\end{figure*}

To solve our highly nonlinear optical flow energy Eq.~\ref{eq:Eng}, we follow a \emph{Nested Fixed Point} based optimization scheme~\cite{Brox} which has been recently used in the state-of-the-art approaches. We define:

{\setlength\abovedisplayskip{-1mm}
\setlength\belowdisplayskip{-1mm}
\begin{eqnarray}
\begin{array}{ll}
B_{x}=\partial_{x}B_{2}(\textbf{x}+\textbf{w})& B_{yy}=\partial_{yy}B_{2}(\textbf{x}+\textbf{w}) \\
B_{y}=\partial_{y}B_{2}(\textbf{x}+\textbf{w}) &
B_{z}=b_{2}(\textbf{x}+\textbf{w})-B_{1}(\textbf{x})\\
B_{xx}=\partial_{xx}B_{2}(\textbf{x}+\textbf{w}) & B_{xz}=\partial_{x}B_{2}(\textbf{x}+\textbf{w})-\partial_{x}B_{1}(\textbf{x})\\
B_{xy}=\partial_{xy}B_2(\textbf{x}+\textbf{w}) & B_{yz}=\partial_{y}B_{2}(\textbf{x}+\textbf{w})-\partial_{y}B_{1}(\textbf{x})
\end{array}\nonumber
\end{eqnarray}
}

We first apply \emph{Euler-Lagrange Equations} onto the energy Eq.~\ref{eq:Eng}. The resulting functional is further minimized within a coarse-to-fine fashion (Algorithm 1). We initialize the flow field $\textbf{w}=(0,0)^T$ on the top-coarsest level; and iteratively update the flow field on the next finer level as $\textbf{w}^{i+1} \approx \textbf{w}^{i}+d\textbf{w}^i$. Here $d\textbf{w}$ denotes increments which is still the nonlinearity of the remaining system. Those are the solutions of

{\setlength\abovedisplayskip{-1mm}
\setlength\belowdisplayskip{-1mm}
 \begin{align}
(\phi')_B^{i}\cdot\{B_{x}^i(B_{z}^i+B_{x}^idu^{i}+B_{y}^idv^{i}) \nonumber \\
+\alpha B_{xx}^i(B_{xz}^i+B_{xx}^idu^{i}+B_{xy}^idv^{i})\nonumber \\
+\alpha B_{xy}^i(B_{yz}^i+B_{xy}^idu^{i}+B_{yy}^idv^{i})\} \nonumber \\
-\gamma(\phi')_S^{i}\cdot\nabla(u^i+du^{i})&=0
\label{eq:fixedK_1}\\
\nonumber\\
(\phi')_B^{i}\cdot\{B_{y}^i(B_{z}^i+B_{x}^idu^{i}+B_{y}^idv^{i}) \nonumber \\ +\alpha b_{yy}^i(B_{yz}^i+B_{xy}^idu^{i}+B_{yy}^idv^{i})\nonumber \\
+\alpha B_{xy}^i(B_{xz}^i+B_{xx}^idu^{i}+B_{xy}^idv^{i})\}\nonumber \\
-\gamma(\phi')_S^{i}\cdot\nabla(v^i+dv^{i})&=0
\label{eq:fixedK_2}
\end{align}
}

where the terms $(\phi')_{B}^{i}$ and $(\phi')_{S}^{i}$ contained $\phi$ provide robustness to flow discontinuity on the object boundary. In addition, $(\phi')_{S}^{i}$ is also regularizer for a gradient constraint in motion space. All of those terms can be detailed as follows:

{\setlength\abovedisplayskip{-1mm}
\setlength\belowdisplayskip{-1mm}
\begin{align}
(\phi')_{B}^{i}&=\phi'\{(B_{z}^i+b_{x}^idu^i+B_{y}^idv^i)^2 \nonumber \\ &+\alpha(B_{xz}^i+v_{xx}^idu^i+v_{xy}^idv^i)^2 \nonumber \\
&+\alpha(B_{yz}^i+B_{xy}^idu^i+B_{yy}^idv^i)^2\} \nonumber \\
(\phi')_{S}^{i}&=\phi'\{\left \|\nabla(u^i+du^i)\right \|^2+\left \|\nabla(v^i+dv^i)\right \|^2\}
\end{align}
}

For the further linearization on the system Eqs.~(\ref{eq:fixedK_1}, \ref{eq:fixedK_2}), please refer to our supplementary document.

\begin{table*}[t]
\centerline{
    \begin{tabular}{|l||c|cc|cc|cc|cc||cc|cc|}
    \hline
        & \multicolumn{1}{c|}{\bfseries{}}
        & \multicolumn{2}{c|}{\bfseries{Urban2}}
        & \multicolumn{2}{c|}{\bfseries{Market2}}\\
        \bfseries{Baselines}
        & \bfseries{Time}
        & \multicolumn{1}{c}{AEE}
        & \multicolumn{1}{c|}{AAE}
        & \multicolumn{1}{c}{AEE}
        & \multicolumn{1}{c|}{AAE}\\ 
        \hline
        Independent Deblurring + Flow EnergyOpt& \textbf{9} &3.66&5.78&6.63&12.46\\
        Independent Deblurring + Blur Matching + Flow EnergyOpt&\textbf{9} &2.07&4.11&5.17&10.60\\
        Per-Level Deblurring + Flow EnergyOpt&29&2.33&4.02&5.59&9.91\\
        Per-Level Deblurring + Blur Matching + Flow EnergyOpt, (\textbf{LMoF})&29&\textbf{1.28}&\textbf{2.99}&\textbf{4.23}&\textbf{8.90}\\
        \hline
    \end{tabular}
    }
    \vspace{2mm}
    \caption{{Quantitative Measure on GT sequences Urban2 and Market2. Four variations of our methods are evaluated along with different deblurring methods (Independent or Per-Level) and Blur Matching strategies (on or off) on the metrics of Average Endpoint Error (AEE), Average Angle Error (AAE) and Average Time Consumption (in second).}}
    \label{tab:eva_self}
\end{table*}

\begin{table*}[t]
\centerline{
    \begin{tabular}{|l||c|cc|cc|cc|cc||cc|cc|}
    \hline
        & \multicolumn{1}{c|}{\bfseries{}}
        & \multicolumn{2}{c|}{\bfseries{Urban2}}
        & \multicolumn{2}{c|}{\bfseries{Market2}}\\
        \bfseries{Baselines}
        & \bfseries{Time}
        & \multicolumn{1}{c}{AEE}
        & \multicolumn{1}{c|}{AAE}
        & \multicolumn{1}{c}{AEE}
        & \multicolumn{1}{c|}{AAE}\\ 
        \hline
        Independent \textbf{Chakrabarti}~\cite{Chakrabarti2016} + BM + FE& 139 &3.69&6.37&6.51&12.21\\
        Per-Level \textbf{Chakrabarti}~\cite{Chakrabarti2016} + BM + FE& 802&2.91&5.83&6.04&11.66\\
        Independent \textbf{Hradi{\v{s}}~\emph{et al.}}~\cite{hradivs2015convolutional} + BM + FE& 60 &4.19&7.84&6.92&12.97\\
        Per-Level \textbf{Hradi{\v{s}}~\emph{et al.}}~\cite{hradivs2015convolutional} + BM + FE& 360 &3.28&7.69&5.64&12.33\\
        Independent \textbf{Xu\&Jia}~\cite{xu2010two} + BM + FE& 63 &2.13&5.44&5.29&9.71\\
        Per-Level \textbf{Xu\&Jia}~\cite{xu2010two} + BM + FE& 363 &3.42&5.96&6.66&9.92\\
        Independent \textbf{Levin~\emph{et al.}}~\cite{levin2011efficient} + BM + FE& 275 &4.47&7.80&7.19&11.71\\
        Per-Level \textbf{Levin~\emph{et al.}}~\cite{levin2011efficient} + BM + FE& 1563 &5.12&7.89&7.91&12.23\\
        \hline
        Independent \textbf{ours} + BM + FE&\textbf{9}&2.07&4.11&5.17&10.60\\
        Per-Level \textbf{ours} + BM + FE, (\textbf{LMoF})&29&\textbf{1.28}&\textbf{2.99}&\textbf{4.23}&\textbf{8.90}\\
        \hline
    \end{tabular}
    }
    \vspace{2mm}
    \caption{{Quantitative Measure on GT sequences Urban2 and Market2. Two of our implementations are compared to eight variations which combine different deblurring baselines (Independent or Per-Level) into our optical flow framework using our blur matching strategy (BM) and energy optimization (FE).}}
    \label{tab:eva_deblur}
\end{table*}

\subsection{Implementation}

In the implementation, we use a customized C++/CUDA version of Caffe~\cite{jia2014caffe} for both the network training and testing. In the training period, we sample 8 directions ($k=8$) for the directional filtering. The training takes around a week for each of iteration in a platform with Intel i7 3.5 GHz and GTX 780 4Gb. Furthermore, we implement the optical flow framework using C++; and construct the image pyramid with a downsampling factor of 0.8. The final system is solved using \emph{Conjugate Gradients} with 60 iterations.

\section{Evaluation}
\label{sec:eva}

In this section, we perform an evaluation by comparing three variations of our proposed approach -- i.e. \emph{LMoF}, \emph{LMoF-Deep} and \emph{LMoF-NoDF} (Sec~\ref{sec:train}) -- to four other famous optical flow methods, i.e. Portz~\emph{et al.}~\cite{Portz}, Li~\emph{et al.}~\cite{moblur_nc}, MDP~\cite{xu2012motion} and Brox~\emph{et al.}~\cite{Brox}. Portz~\emph{et al.}'s approach introduce the uniform blurry parameterizations and provides sharp image alignment for both the camera-shake and object blur cases. Li~\emph{et al.}'s method brings the directional filtering to give the recent state-of-the-art precision for the camera-shake blur. MDP is currently one of top method according to Middleburry benchmark~\cite{Middlebury} while Brox~\emph{et al.}'s show the similar optimization scheme to the proposed method. We use the default parameter setting for all baselines.

In the following subsections, we evaluate our method on a synthetic GT dataset, as well as real-world sequences respectively.

\subsection{Customized Benchmark}

It is difficult to quantitatively evaluate the optical flow from real-world blurry scenes which may lead to the ambiguous matching issue. Portz~\emph{et al.} propose a synthetic benchmark that gives blurry object motion within a blur-free background but lack of camera-shake blur. Furthermore, by carefully sampling the useful correspondences, Li~\emph{et al.}~\cite{moBlur} synthesize a benchmark for camera-shake blurred scenes by convoluting selective blur kernels onto a customized subset of the famous Middleburry dataset. Such benchmark is challenging as it contains many small details that can be easily destroyed by blur.

In this evaluation, we bring more challenges. As shown in Fig.~\ref{fig:gt}, we synthesize two additional GT sequences applying Li~\emph{et al.}'s GT methodology onto selective Sintel~\cite{Sintel} sequences (Market3 and Bamboo1, downsampling to $615 \times 262$ pix.). Such extra benchmark is supposed to give more difficulties e.g. mixed blur, large displacement and illumination changes, etc.

{Table~\ref{tab:eva} illustrates the quantitative comparison of our methods (three implementations) against the other baselines. Our LMoF-Deep yields the best Average Endpoint Error (AEE) precision for all the sequences. It also competitively ranks the second best Average Angular Error (AAE) for the Market2, and offers the top AAE measure for all other trials. Li~\emph{et al.}'s is the state-of-the-art approach in the community and provides very competitive precision measure comparing to LMoF -- a fast version of our method. Their approach results in the second best AEE accuracy for the Grove2, Hydrangea, Rub.Whale and Bamboo1, as well as the third best AEE measure on the Urban2. Our other implementations of LMoF and LMoF-NoDF also outperform the baselines Portz~\emph{et al.}'s, Brox~\emph{et al.}'s and MDP for most of the trials. All our implementations show reasonable speed in the experiments. Note that most baselines give relevantly larger errors ($>$ 3 pixel AEE and $>$ 6 degrees AAE) on the Market2 because the sequence contains additional difficulties e.g. invariant blur (motion blur and camera blur), large displacements and noise.}

\begin{figure*}[t]
\centerline{
\includegraphics[width=0.95\linewidth]{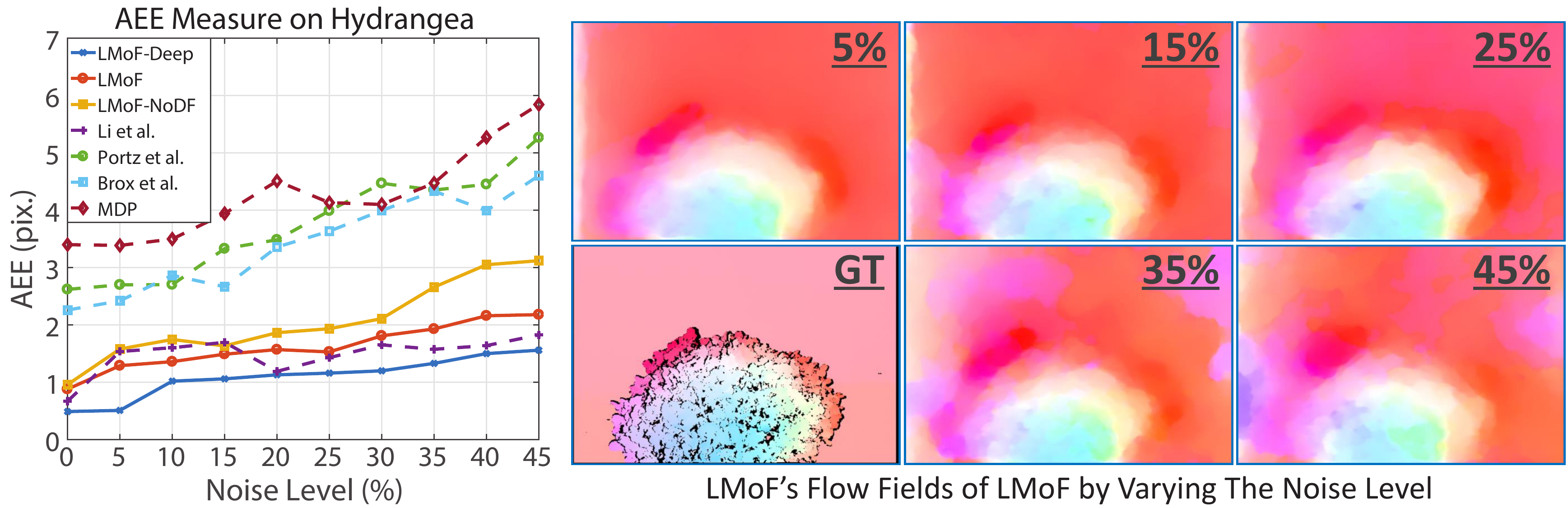}
}
\caption{AEE Measure on Hydrangea by Ramping Up the Noise Distribution. \textbf{Left}: the numerical analysis by varying noise level. \textbf{Right}: the visualizations on flow fields.}
\label{fig:noiseErr}
\end{figure*}

\begin{figure*}[t]
\centerline{
\includegraphics[width=0.98\linewidth]{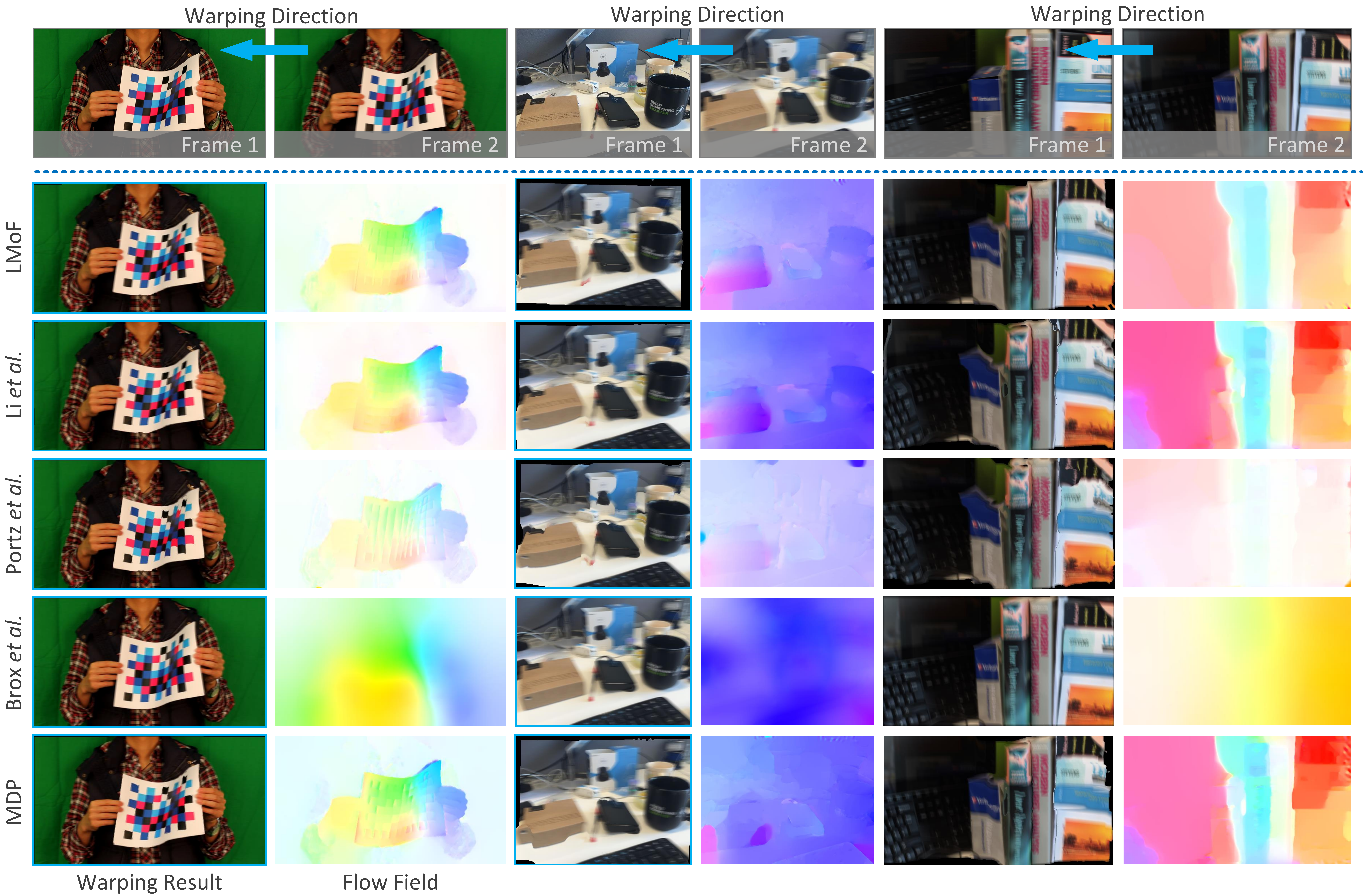}
}
\caption{Visual Comparison on Real-world Sequences of Chessboard, Desktop and Books. \textbf{First Row:} two input frames. For the rest, from left to right, the \textbf{Second}, \textbf{Forth} and \textbf{Last Columns} are the flow fields of each baselines. The \textbf{First} \textbf{Third} and \textbf{Third Columns} are the warping results using each baseline flow fields.}
\label{fig:realErr}
\end{figure*}

{Table~\ref{tab:eva} also demonstrates the advantage of our method comparing to two neural network based optical flow approaches, i.e. FlowNet~\cite{dosovitskiy2015flownet} (FlowNetS and FlowNetC) and Teney\&Hebert~\cite{teney2016learning} which provide an end-to-end process to recover optical flow from input images. We observe that both implementations of FlowNet (FlowNetS and FlowNetC) yield large error for the small motion scenes (Grove2, Hydragnea, Rub.Whale and Urban2) while they give relatively higher accuracy for the large motion cases i.e. Bamboo1 (2.00 pixel AEE) and Market2 (4.01 pixel AEE). Furthermore,  Teney\&Hebert encodes a hidden coarse-to-fine optimizer within the network. With this advantage, they give improved results for the small motion scenes and outperform the traditional approaches Brox~\emph{et al.} and MDP in most of trials. However, our methods produce the top precision measure for all the sequences except Matket2 (FlowNetS, 8.18 degrees AAE).}

Fig.~\ref{fig:synErr} visualizes the AEE errors of all the baselines on Bamboo1 and Market2. Our methods yield less details loss and clearer object boundaries in overall. Here Brox~\emph{et al.}'s overly smooth the object details of the scene. And MDP leads to extra errors because their feature detection and matching process is compromised by the blur, and even brings error into the final energy. We observe that all the baselines result in large errors on left area of the Market2 because the object there is moving quickly and leads to extra motion blur. Such invariant blur cannot be solved by any of our baselines, as well as is out of this paper's scope.

{Moreover, Table~\ref{tab:eva_self} shows the quantitative analysis given different deblurring strategies on our proposed approach. Here \textbf{Per-level} denotes the deblurring strategy used in our implementations. For each level of image pyramid (coarse-to-fine), our deblurring network stacks the blur image and the latent one propagated from previous level in order to compute the optimized latent image. This latent image is then propagated to the next level. Hence, on the final level, our deblurring network runs $N$ (the number of network iterations, Fig.~\ref{fig:nnet}) $\times$ $M$ (the number of levels of image pyramid) network iterations on each of input images. However, \textbf{Independent} deblurring presents the process where our deblurring network and optical flow optimization are treated as two independent steps. In this case, deblurring network runs only once on the full resolution images. We observe that our \textbf{Per-level} is able to significantly improve the precision on both small (Urban2) and large motion (Market2) scenes but use longer time (29s vs. 9s) for computation. The quantitative analysis also illustrates that the \textbf{Blur Matching}(see Sec,~\ref{sec:robustEng}) can also improve the final results.}

{In Table~\ref{tab:eva_deblur}, we further evaluate how our deblurring network contributes to the final optical flow estimation. To highlight our advantage, we propose eight variations by replacing our deblurring network with four selected deblurring approaches in either \textbf{Independent} or \textbf{Per-level} fashion. Hradi{\v{s}}~\emph{et al.}~\cite{hradivs2015convolutional} and Chakrabarti~\cite{Chakrabarti2016} are neural network based approaches. The former gives high quality deblurring result on fine details e.g. text and license plate; while the latter achieves the state-of-the-art for the general real-world scenes. Levin~\emph{et al.}~\cite{levin2011efficient} and Xu\&Jia~\cite{xu2010two} are non-learning methods. The former is one of the most popular approaches in practice; while the latter shows high performance on the noisy image. Please note that we adopt the default and fixed parameter setting through all the trials.}

{It is observed that our method (both Independent and Per-level) yields the best precision measure for both trials while they are also much faster than any other baselines. We also observe that Hradi{\v{s}}~\emph{et al.}~\cite{hradivs2015convolutional} and Chakrabarti~\cite{Chakrabarti2016} provide improved error measure when they are applied by a \textbf{Per-level} deblurring strategy. However, Levin~\emph{et al.}~\cite{levin2011efficient} and Xu\&Jia~\cite{xu2010two} result in relevantly higher accuracy when they are performed as an \textbf{Independent} process. Our optimization framework (FE) contains a coarse-to-fine image pyramid in a top-down fashion. In the Per-level deblurring strategy, the baseline has to be performed on different resolutions (different levels of image pyramid) of the image. It is difficult for the non-learning methods to adapt to different resolutions without manually tuning the parameters. This issue may bring extra errors. However, the neural network based approaches are supposed to improve this issue if the training data is sufficient to cover different sizes of blur kernels.}

Using the sequence Hydrangea, Fig.~\ref{fig:noiseErr} quantizes and visualizes the effects by ramping up the distribution of the noise. By increasing the distribution of noise, all of our baselines give more errors in overall. The AEE of our implementations are still on the relevantly reasonable level ($>$ 3.2 pix. AEE) while the errors of Brox~\emph{et al.}, Portz~\emph{et al.} and MDP are climbing up quickly. Given the largest noise level ($45\%$), our LMoF-Deep gives the best precision (1.61 pix. AEE). And the Li~\emph{et al.}'s yields a very competitive measure (1.88 pix. AEE) while LMoF gives 2.13 pix. AEE. The robustness of those three approaches against the noise benefits from the directional filtering which efficiently removes the noise but preserves the useful information~\cite{moBlur}.

{Within this evaluation, we compare our proposed approach to recently popular Li~\emph{et al.}~\cite{moblur_nc} which uses ground truth camera motion to regularize the optical flow estimation. They give good precision on real-world blurry footages but additional hardware and difficult calibration are strictly required. They also have to tune parameters carefully for various scenes. Our method models the optical flow from blurry footage using convolutional neural network. This is an end-to-end unsupervised approach which does not need any manual parameter tuning or additional information/hardware. It is able to provide rapid computation and adapt to various image resolutions and kernel sizes. In our quantitative analysis (Table~\ref{tab:eva}), our method produces more than $30\%$ AEE improvement and $10\%-25\%$ faster comparing to Li~\emph{et al.}~\cite{moblur_nc}.}

\subsection{Real-world Scenes with Camera-shake Blur}

To illustrate the feasibility of our method, we qualitatively compare our approach to other baselines on the real-world sequences. As shown in Fig.~\ref{fig:realErr}, from left to right, there are sequences Chessboard, Desktop and Books. Chessboard contains real-world photometric effects of nonrigid deformations and small occlusions while the Desktop represents the large camera motion and some featureless regions. Books give large displacement and out-of-plane rotation. We observe that our methods give the sharper flow on object boundaries, as well as shape preservation in the image warping.

\section{Conclusion}

In this paper, we investigate the problem for recovering optical flow from a camera-shake video footage. We first propose a novel CNNs architecture for video frame deblurring using an extra directional similarity and filtering layer. In practice, such learnable filters are able to adoptively preserve the directional blur information without the pre-knowledge of the camera motion. We then highlight the benefits of the Per-level integration of our network into an iterative optical flow framework. The evaluation demonstrates our hybrid framework gives the overall competitive precision and higher performance in runtime.

The limitations of our method may lie in the presence of mixed blur, globally invariant blur and spatial noise. Such difficulties could be improved by using more comprehensive training data.

\section*{Acknowledgements}
This work was partially conducted when Wenbin Li was affiliated to UCL Department of Computer Science and University of Bath. We thank Gabriel Brostow and the UCL PRISM Group for their helpful comments. The authors are partially supported by Centre for the Analysis of Motion, Entertainment Research and Applications (CAMERA) EP/M023281/1; and EPSRC projects EP/K023578/1 and EP/K02339X/1.



\bibliographystyle{elsarticle-num}
\bibliography{bib}





\end{document}